\documentclass[letterpaper,11pt,onecolumn]{IEEEtran} 
\IEEEoverridecommandlockouts                              

\usepackage{amsfonts, amssymb, amsmath}
\usepackage{epsfig}
\usepackage{epstopdf}
\usepackage{theorem}
\usepackage{textcomp}
\usepackage{tipa}
\usepackage{multirow}
\usepackage{tabularx}
\usepackage[noadjust]{cite}
\makeatletter
\let\NAT@parse\undefined
\makeatletter
\usepackage{natbib}

\let\chapter\section 
\usepackage[linesnumbered,ruled]{algorithm2e}
\SetKwInput{KwIn}{Initial condition}
\SetKwInput{KwResult}{Outcome}

\newtheorem{theorem}{Theorem}

\newtheorem{problem}{Problem}
\newtheorem{objective}{Objective}

\newcommand{\qed}{\hfill $\Box$\\}
\newcommand{\rqed}{\hfill $\triangle$\\}

\def\remark{\noindent\textbf{Remark. }}
\def\mpp{\textrm{MPP}}
\def\pmpp{\textrm{PMPP}}
\def\sat{{\sc 3-SAT}}
\def\psat{{\sc MP3SAT}}
\def\pmtmpp{{\sc MMPmpp}}
\def\pttmpp{{\sc MTTPmpp}}
\def\pmdmpp{{\sc MMDPmpp}}
\def\ptdmpp{{\sc MTDPmpp}}

\author{Jingjin~Yu$^{1}$ 
\thanks{$^{1} $J. Yu is with the Department of Computer Science, Rutgers University at New Brunswick, Piscataway, New Jersey, USA 08854. E-mail: {\tt\small jingjin.yu@rutgers.edu}. %
This work is supported by a startup fund from Rutgers Unviersity.}
}
\title{Intractability of Optimal Multi-Robot Path Planning on Planar Graphs}

\begin{document}
\maketitle
\begin{abstract}We study the computational complexity of optimally solving multi-robot path planning problems on planar graphs. For four common time- and distance-based objectives, we show that the associated path optimization problems for multiple robots are all NP-complete, even when the underlying graph is planar. Establishing the computational intractability of optimal multi-robot path planning problems on planar graphs has important practical implications. In particular, our result suggests the preferred approach toward solving such problems, when the number of robots is large, is to augment the planar environment to reduce the sharing of paths among robots traveling in opposite directions on those paths. Indeed, such efficiency boosting structures, such as highways and elevated intersections, are ubiquitous in robotics and transportation applications. \end{abstract}
\vspace{2mm}
\begin{IEEEkeywords}NP-hardness, multi-robot path planning, planar graphs, transportation networks, boolean satisfiability problems.\end{IEEEkeywords}

\section{Introduction}\label{section:introduction}
\IEEEPARstart{I}{n} this paper, we establish the computational complexity of optimally solving multi-robot path planning problems on planar graphs. In such a problem, a set of robots is initially located on the vertices of a connected planar graph. The task is to move the robots to a different set of vertices (each robot must reach a specific goal) of equal cardinality, in an optimal manner and collision free. Here, we look at four common optimality objectives with two focused on time optimality and two focused on distance optimality. For each of these objectives, we prove that the associated multi-robot optimal path planning problem is NP-hard. Since the decision versions of these problems are in NP, they are NP-complete. 

{\bf Motivation and related work.} Our study is primarily motivated by the emerging trend in deploying multi-robot systems to automatically carry out complex tasks \cite{Boh04,GroKelKumPap06,WurDanMou08,KneLayRomRus13}. In these and many other applications, it is often highly desirable to plan for and control the robots so that they could optimally reach their respective target locations according to some metric, such as time- or distance-based measures. For example, for the Kiva Systems' robots operating in a warehouse \cite{WurDanMou08}, the end goal is to put together orders for shipment as quickly as possible. This translates to a time-optimal travel requirement for the robots. A natural question then arises: can we solve these optimization problems for large problem instances efficiently, for example in low polynomial time? Knowing the answer to this question has important practical relevance. If the answer is positive, then the accompanying algorithmic solution will boost operating efficiency. On the other hand, if the answer is negative, one should perhaps look beyond algorithmic solutions in solving such problems, especially when the problem instance becomes larger and larger. For example, engineering the environment ({\em i.e.}, the underlying graph structure in our problem) has long been applied in transportation system design, which frequently employs highways and elevated intersections to improve traffic throughput. Similar  effort for simplifying problem solving is also observed in robotics applications \cite{RooRaf11}, in which ``highways'' are designed to speed up the operation of Kiva Systems' mobile robots. As our study establishes a firm ``no'' answer to this complexity question, our results offer theoretical justification for adopting environment engineering as a preferred approach for attacking optimal multi-robot path planning and closely related transportation problems, through the computation angle. 

The study of multi-robot path planning in a graph-theoretic setting originates from the mathematical study of the 15-puzzle \cite{Sto1879, Wil74}, popularized by Sam Loyd \cite{Loy59}. A generalized version of the problem, called the {\em pebble motion} problem, is proposed in \cite{KorMilSpi84}, for which a cubic time algorithm is provided to find a feasible solution. Although only a single robot is allowed to move in a time step in the formulation given in \cite{KorMilSpi84}, it is conceivable that multiple robots could be allowed to move simultaneously as long as no collision is incurred between any two robots. We denote this version of the problem, allowing concurrent collision-free robot movements, as \mpp, which stands for Multi-robot Path Planning on graphs. 

Given the immediate applicability of \mpp\, to domains such as computer games and robotics, many algorithms have been proposed to solve it according to some optimality criteria \cite{Rya08,StaKor11,WagChoC11,Sur12}, of which most are time- or distance-based. Whereas great progress has been made, no algorithm is known to optimally solve these problems in polynomial time. This suggests that the problem may be computationally intractable. Indeed, the intractability of optimally solving \mpp\, and related problems has been established  \cite{Gol84,RatWar90,Sur10,YuLav13AAAI}. However, to the best of our knowledge, no existing work addresses a planar setting while allowing general concurrent movements from multiple robots, which directly mirrors environments such as warehouses for Kiva's robots or road networks for automobiles. We note that, the formulation in \cite{RatWar90} works with grids which are planar. However, in \cite{RatWar90}, only a single robot is allowed to move to the single empty (swap) vertex at each time step ({\em i.e.}, it works with a restricted \mpp\, formulation); no concurrent robot movement is permitted. In contrast, we allow synchronous robot motion on graphs containing an arbitrary number of empty vertices, including the case with no swap vertex. 

The computational complexity of multi-robot path planning in two-dimensional continuous domains has also been extensively studied. The feasibility problem of translating rectangular blocks in a rectangular workspace has long been established as PSPACE-hard \cite{HopSchSha84}. The problem becomes strongly NP-hard when the blocks become discs with varying radii residing in a simple polygon \cite{SpiYak84}. Recently, it is shown that the problem is PSPACE-hard even for unlabeled squares \cite{SolHal15}, the proof of which uses the non-deterministic constraint logic model \cite{HeaDem05}. We mention that, the associated hardness proofs on feasibility from these work do not readily extend to optimal \mpp\, for two reasons: {\em (i)} these proofs rely on geometric arguments, and {\em (ii)} the feasibility of graph-based \mpp\, is in P \cite{YuRus14WAFR}.  

{\bf Contributions.} Denoting the planar version of the \mpp\, problem as \pmpp, the main contribution of our work is showing rigorously that computing many time- and distance-optimal solutions for \pmpp\, is NP-hard. For time-optimality, we further prove that such intractability persists even when there are only two groups of robots such that the robots are indistinguishable within each group. Moreover, due to the obvious NP membership of these problems, the decision versions of these problems are NP-complete. 

From a practical standpoint, because \pmpp\, relates to real world path planning and transportation problems in which robots or vehicles move on some form of road networks embedded in a two-dimensional plane, our NP-hardness result convincingly demonstrates that such problems are computationally demanding to optimally solve. This suggests, for optimally coordinating the movements of a large number of robots (or vehicles) in a planar setting, environment augmentation should be explored in addition to algorithmic improvements. Indeed, in practice, we observe that engineered solutions such as highways and elevated intersections are widely adopted, which can be readily justified with our findings.  

{\bf Organization.} The rest of the paper is organized as follows. We define the \pmpp\, and optimal versions of the problem In Section~\ref{section:formulation}. In Section~\ref{section:time}, we prove the NP-hardness of the time-optimal formulations and also show the intractability remains when there are only two groups of robots. In Section~\ref{section:distance}, we show the intractability of distance-optimal formulations. We conclude the paper in Section~\ref{section:conclusion}. 

\section{Multi-robot Path Planning on Planar Graphs: Basic and Optimal Formulations}\label{section:formulation}
\subsection{The multi-robot path planning problem}
Let $G = (V, E)$ be a {\em connected}, {\em undirected}, {\em simple}, {\em planar graph}, with $V = \{v_i\}$ being the vertex set and $E = \{(v_i, v_j)\}$ the edge set with unit edge length. A graph is {\em planar} if its edges can be embedded in the two-dimensional plane without any two edges crossing each other. Let $R = \{r_1, \ldots, r_n\}$ be a set of $n$ robots. A robot may stay still or move at unit speed along edges of $G$. A {\em configuration} of the robots is an injective map from $R$ to $V$. The start and goal configurations of the robots are denoted as $x_I$ and $x_G$, respectively.

A {\em scheduled path} is a map $p_i: \mathbb Z^+ \to V$, in which $\mathbb Z^+ := \mathbb N \cup \{0\}$. A path set $P = \{p_1, \ldots, p_n\}$ is {\em feasible} if it takes the robots to their respective goals. More formally, $P = \{p_i\}$ must satisfy the following properties: 1) $p_i(0) = x_I(r_i)$. 2) For each $i$, there exists a smallest $t_i \in \mathbb Z^+$ such that $p_i(t_i) = x_G(r_i)$. 3) For any $t \ge t_i$, $p_i(t) \equiv x_G(r_i)$. 4) For any $0 \le t < t_i$, $(p_i(t), p_i(t+1)) \in E$ or $p_i(t) = p_i(t+1)$ (if $p_i(t) = p_i(t+1)$, robot $r_i$ stays at vertex $p_i(t)$ between the time steps $t$ and $t+1$). We say that two paths $p_i, p_{j}$ are in {\em collision} if there exists $t \in \mathbb Z^+$ such that $p_i(t) = p_{j}(t)$ (meet-collision) or $(p_i(t), p_i(t+1)) = (p_j(t+1), p_j(t))$ (head-on-collision)\footnote{We assume that the graph $G$ only allows ``meet'' or ``head-on'' collisions. For example, a (arbitrary dimensional) grid with unit edge distance is such a graph for robots with radii of no more than $\sqrt{2}/4$.}. 

\begin{problem}[Planar Multi-Robot Path Planning ($\pmpp$)]\label{mpp} Given $(G, R, x_I, x_G)$, find a feasible path set $P = \{p_1, \ldots, p_n\}$ such that no two paths $p_i, p_j \in P$ are in collision. 
\end{problem}

\remark We point out that the feasibility of \pmpp\, can be decided in linear time and a feasible \pmpp\, instance can be solved in polynomial time \cite{YuRus14WAFR}. Alternatively, if there is a single robot, optimal path planning is also in P \cite{dijkstra1959note}. ~\rqed

\subsection{Optimal Formulations}
Let $P = \{p_1, \ldots, p_n\}$ be an arbitrary feasible solution to a fixed $\pmpp$ instance. For a path $p_i \in P$, $len(p_i)$ denotes the length of the path $p_i$, which is increased by one each time when the robot $r_i$ passes an edge. A robot, following $p_i$, may visit the same edge multiple times. Recall that $t_i$ denotes the arrival time of robot $r_i$. In optimizing over the solutions to $\pmpp$, we examine four common objectives with two focusing on time optimality and two focusing on distance optimality. Each objective is stated together with the corresponding decision problem required for stating NP-completeness results.  

\begin{objective}[Min Total Time]\label{ott} Compute a path set $P$ that minimizes 
\begin{displaymath}
\sum_{i = 1}^nt_i.
\end{displaymath}
\end{objective}

\vspace{1mm}
\noindent \pttmpp\, (Min Total Time $\pmpp$) \\
\noindent Instance: An instance of $\pmpp$, and $k \in \mathbb Z$.\\
\noindent Question: Is there a solution path set $P$ with a total arrival time no more than $k$? 

\begin{objective}[Min Makespan]\label{omakespan} Compute a path set $P$ that minimizes 
\begin{displaymath}
\max_{1 \le i \le n}t_i.
\end{displaymath}
\end{objective}

\vspace{1mm}
\noindent \pmtmpp\, (Min Makespan $\pmpp$) \\
\noindent Instance: An instance of $\pmpp$, and $k \in \mathbb Z$.\\
\noindent Question: Is there a solution path set $P$ with a makespan no more than $k$? 

\begin{objective}[Min Total Distance]\label{otd}Compute a path set $P$ that minimizes 
\begin{displaymath}
\sum_{i = 1}^n len(p_i).
\end{displaymath}
\end{objective}

\vspace{1mm}
\noindent \ptdmpp\, (Min Total Distance $\pmpp$) \\
\noindent Instance: An instance of $\pmpp$, and $k \in \mathbb Z$.\\
\noindent Question: Is there a solution path set $P$ with a total path distance no more than $k$? 

\begin{objective}[Min Max Distance]\label{od}Compute a path set $P$ that minimizes 
\begin{displaymath}
\max_{1 \le i \le n}len(p_i).
\end{displaymath}
\end{objective}

\vspace{1mm}
\noindent \pmdmpp\, (Min Max Distance $\pmpp$) \\
\noindent Instance: An instance of $\pmpp$, and $k \in \mathbb Z$.\\
\noindent Question: Is there a solution path set $P$ in which every path has a distance no more than $k$? 
\vspace{1mm}

\begin{figure}[ht!]
\begin{center}
    \includegraphics[width=0.26\textwidth]{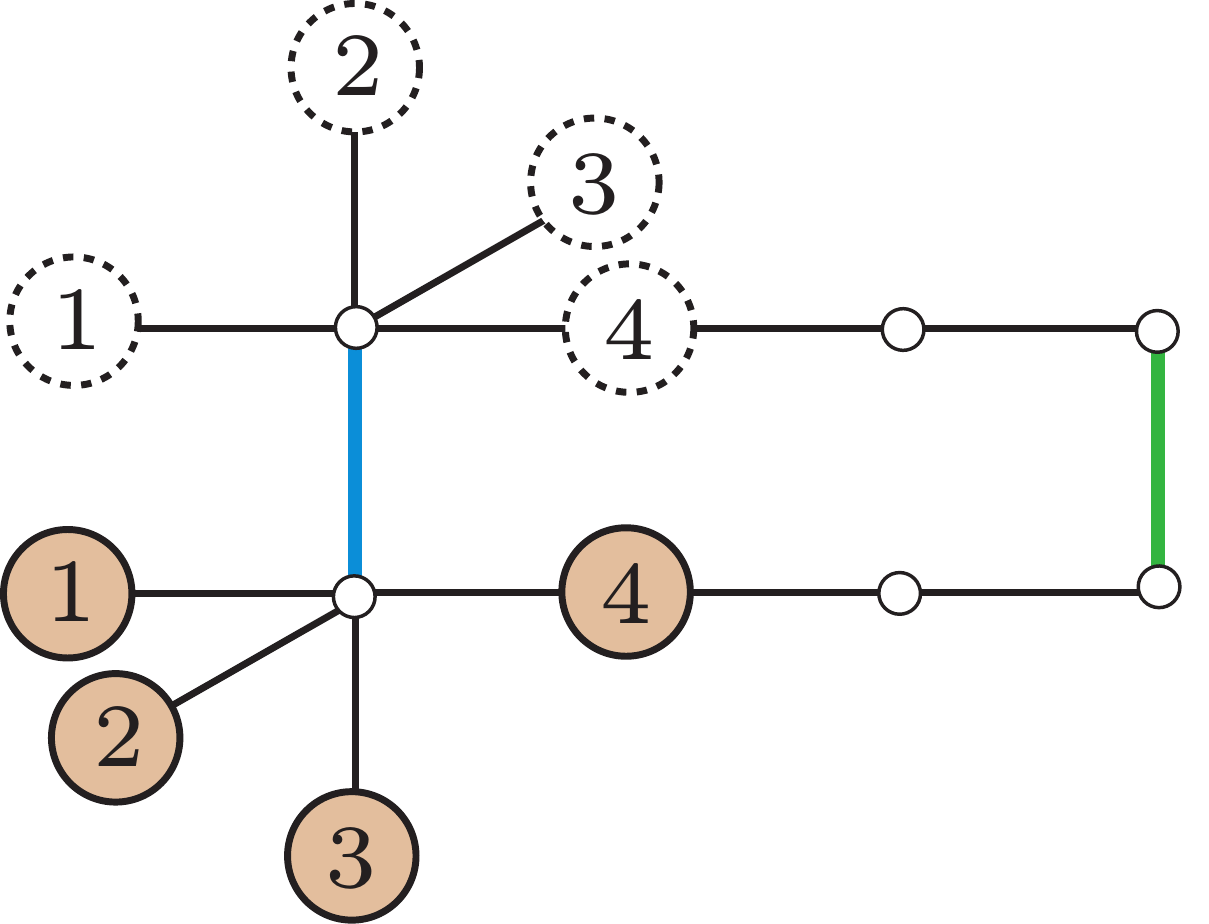} 
\end{center}
\caption{\label{figure:tt-td} An instance of a planar multi-robot path planning problem. The labeled, shaded discs are the start locations of the robots and the labeled, unshaded discs are the goal locations.}
\end{figure}

We illustrate what these objectives achieve using the example shown in Fig.~\ref{figure:tt-td}. 
The individual costs incurred by the robots are listed in Table~\ref{table:example}. In a minimum total time solution, robots $1$, $2$, and $3$ must sequentially travel through the bold (blue) vertical edge on the left side of Fig.~\ref{figure:tt-td}, yielding arrival times of $3, 4$, and $5$, respectively. Robot $4$ should instead travel through the bold (green) vertical edge on the right, yielding an arrival time of $5$. This solution is also a min-max time solution for this particular example. For distance-optimal solutions, all robots should go through the bold vertical edge on the left, yielding a per-robot distance of $3$. In general, however, an arbitrary pair of these four objectives creates a Pareto front (see {\em e.g.}, \cite{YuLav13AAAI}). That is, it is not always possible to simultaneously optimize any two of these four objectives. 

\vspace{1mm}
\remark It is important to note the relationship between \pmpp\, and \mpp, the non-planar formulation with the only distinction being that $G$ is not required to be planar. The computational intractability of \mpp\, has been established in \cite{YuLav13AAAI} for several objectives. We note that if an optimal \pmpp\, formulation is NP-hard, then the corresponding \mpp\, formulation is also NP-hard. This is true because \mpp\, contains \pmpp. The NP-hardness of an optimal \mpp\, problem, however, does not directly imply the NP-hardness of the corresponding \pmpp\, problem because a non-planar graph cannot be readily turned into a planar graph. ~\rqed
\begin{table}[h]
\begin{center}
	\caption{\label{table:example}Minimum required time and distance for the robots.}
	\begin{tabularx}{\columnwidth}{ccXXXX}
   \hline\hline
	 Robot  && 1 & 2 & 3 & 4 \\
	 \hline
	 Total time / max time && 3 & 4 & 5 & 5 \\
	 \hline
	 Total distance / max distance && 3 & 3  & 3 & 3 \\
	 \hline\hline
	 \end{tabularx}
\end{center}
\end{table}

\section{Intractability of Time-Optimal Formulations}\label{section:time}

To show intractability, our general strategy is a reduction from a special \sat\, problem \cite{GarJoh79} called the {\em monotone planar} \sat\, \cite{BerKho10}, which we denote as \psat. Starting from an arbitrary \psat\, instance, we construct a corresponding optimal \pmpp\, instance (for each of the four objectives) in polynomial time.  Our \pmpp\, instances are constructed in two phases. In the first phase, we provide a high-level, skeleton graph structure containing {\em directed paths}. In the second phase, we fill in the details on how to {\em simulate} these directed paths in a \pmpp\, instance (as directed paths are not allowed in \pmpp). We then show the \psat\, instance is satisfiable if and only if the \pmpp\, instance has a certain optimal solution. Because \psat\, is NP-hard \cite{BerKho10}, this implies that the various optimal \pmpp\, problems are also NP-hard. 

Our main goal in this section is proving the intractability of computing time-optimal solutions for \pmpp, {\em i.e},

\begin{theorem}\label{t:time-npc} \pttmpp\, and \pmtmpp\, are both NP-hard.\end{theorem}

As the complete proof of Theorem~\ref{t:time-npc} is involved, we begin with a sketch including the necessary preliminaries. A more rigorous proof then follows. 

\subsection{Reducing \psat\, to \pmtmpp: A Sketch}
\begin{figure}[ht!]
\begin{center}
\includegraphics[width=0.60\textwidth]{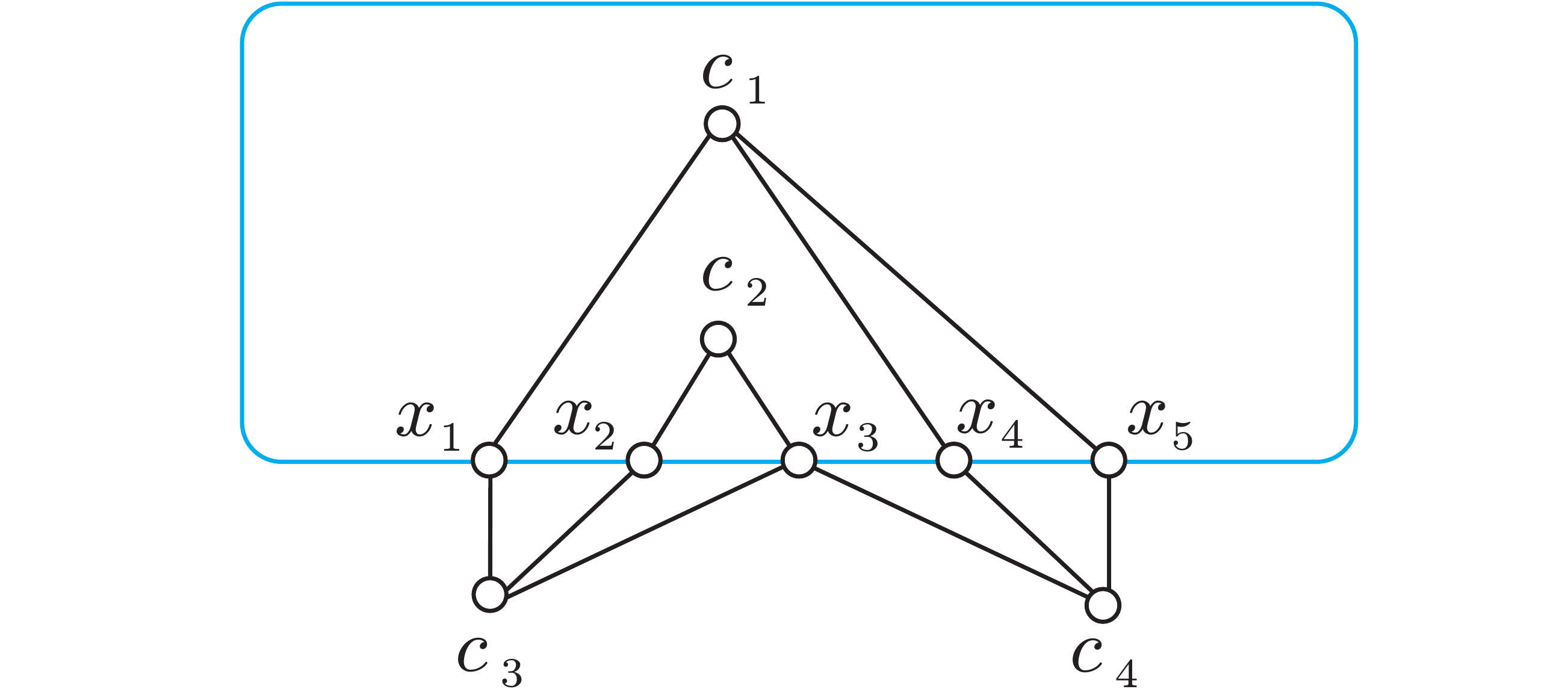} 
\end{center}
\caption{\label{figure:p3sat}The {\em monotone planar} \sat\, instance $(\{x_1, \ldots, x_5\}, \{c_1 = x_1 \vee x_4 \vee x_5, c_2 = x_2 \vee x_3, c_3 = \neg x_1 \vee \neg x_2 \neg x_3, c_4 = \neg x_3 \vee \neg x_4 \neg x_5\}))$ represented as a planar graph.}
\end{figure}

We first introduce \psat, an instance of which has the following structure. There are $n$ variables, $x_1, \ldots, x_n$, and $m$ disjunctive clauses, $c_1, \ldots, c_m$. Each clause $c_j$ contains up to three literals that can either be all non-negated or all negated. A clause with only non-negated (resp., negated) literals is called a {\em positive} (resp., {\em negative}) clause. This is the {\em monotone} element of the \psat\, problem. To describe the {\em planarity} element, we construct a graph based on a monotone \sat\, instance. Given the monotone instance $(\{x_1, \ldots, x_5\}, \{c_1 = x_1 \vee x_4 \vee x_5, c_2 = x_2 \vee x_3, c_3 = \neg x_1 \vee \neg x_2 \vee \neg x_3, c_4 = \neg x_3 \vee \neg x_4 \vee \neg x_5\})$ as an example (see Fig.~\ref{figure:p3sat}), we do the following: 
\begin{enumerate}
\item Add a vertex for each variable $x_i$ and each clause $c_j$,
\item Add an edge between two consecutive variable vertices $x_i$ and $x_{(i+1) \textrm{ mod } n}$ (the blue cycle of variables in Fig.~\ref{figure:p3sat}),
\item For a variable $x_i$ and a clause $c_j$, if $c_j$ has $x_i$ or $\neg x_i$ as a literal, add an edge between vertex $x_i$ and vertex $c_j$. 
\end{enumerate}

The planar element of \psat\, requires that the graph constructed following these rules is planar. In particular, this implies that the positive clauses and the negative clauses are separated by the circle formed by the edges between variable vertices. \psat\, is NP-hard \cite{BerKho10}. 

From the planar structure, we assign each clause a {\em nesting level} that is defined recursively. In the planar graph ({\em e.g.}, Fig.~\ref{figure:p3sat}), if a clause vertex shares a face (a connected 2D region enclosed by edges) with the edge $(x_n, x_1)$, then it is assigned a nesting level of $0$. Otherwise, a clause vertex is assigned a nesting level of $\ell$ if there is a minimum of $\ell - 1$ faces between the clause vertex and the edge $(x_n, x_1)$, without crossing any other $(x_i, x_{i+1})$ edge. In our example, $c_1$, $c_3$, and $c_4$ all have a nesting level of $0$ and $c_2$ is at level $1$. We say that $c_{j_1}$ is {\em directly nested} in $c_{j_2}$ when the nesting level of $c_{j_1}$ and $c_{j_2}$ differ by one, and $c_{j_1}$ is separated from $(x_n, x_1)$ by edges containing $c_{j_2}$. In our example, $c_2$ is directly nested in $c_1$. 

\begin{figure}[ht]
\begin{center}
\includegraphics[width=0.60\textwidth]{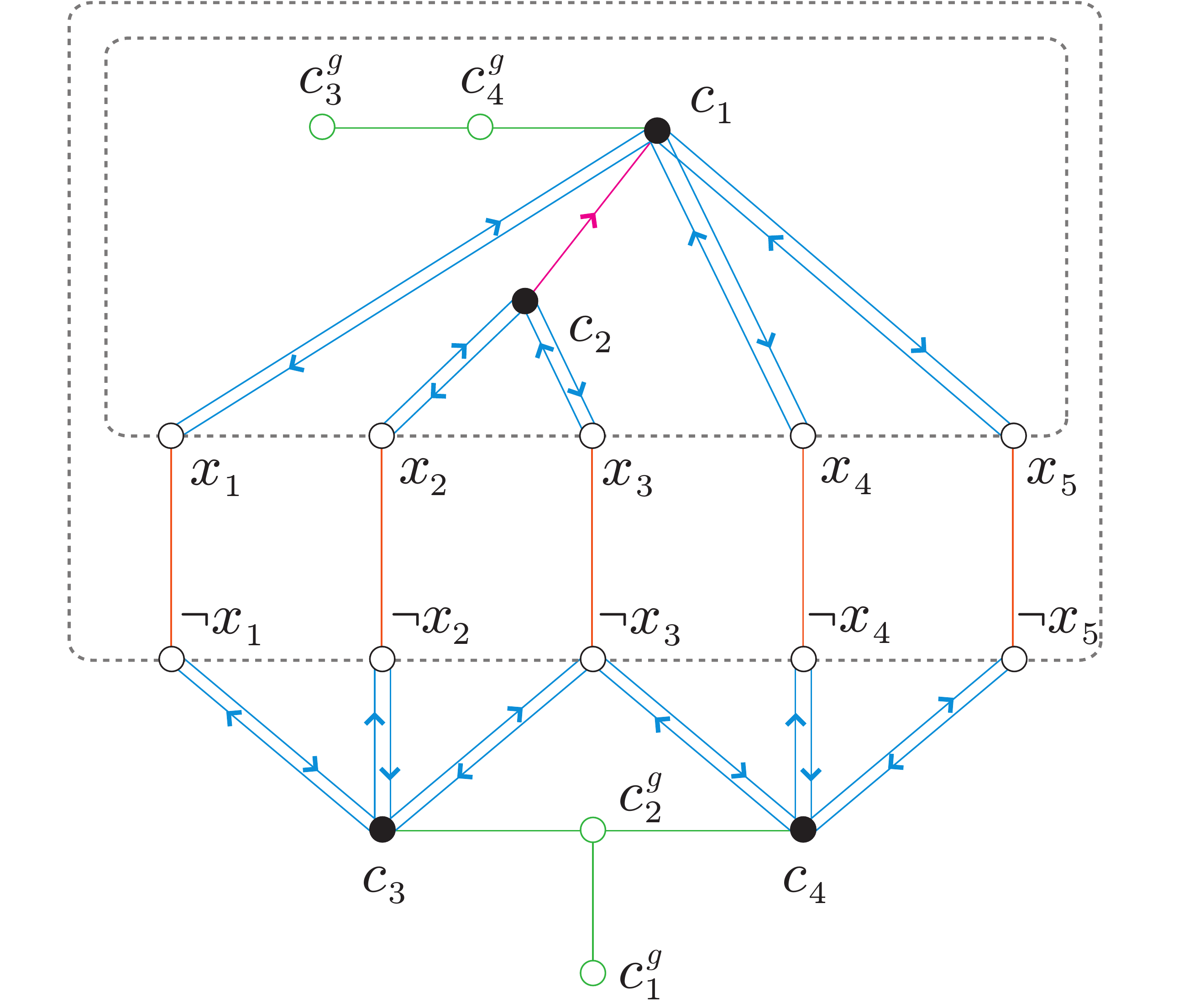} 
\end{center}
\caption{\label{figure:sketch} The skeleton of the \pmtmpp\, instance reduced from the \psat\, instance $(\{x_1, \ldots, x_5\}, \{c_1 = x_1 \vee x_4 \vee x_5, c_2 = x_2 \vee x_3, c_3 = \neg x_1 \vee \neg x_2 \neg x_3, c_4 = \neg x_3 \vee \neg x_4 \neg x_5\}))$, as visualized in Fig.~\ref{figure:p3sat}.}
\end{figure}

After introducing \psat, we construct a \pmtmpp\, instance from a \psat\, instance, using the \psat\, instance from Fig.~\ref{figure:p3sat} as an example. Here, we sketch the construction and the associated reduction proof. The skeleton of the converted \pmtmpp\, instance is given in Fig.~\ref{figure:sketch}. The key idea behind the reduction is to create {\em clause robots} and force unidirectional travel of these robots through {\em variable channels} (the orange paths between $x_i$ and $\neg x_i$). For a clause $c_j$, we denote the corresponding robot as $r_{c_j}$. In comparison to Fig.~\ref{figure:p3sat}, each variable 
vertex is split into an edge and we delete the edges between consecutive variables ({\em i.e.}, the edges $(x_i, x_{(i+1) \textrm{ mod } n})$ are removed). The edges between variable vertices and clause vertices are split into ``directional paths'' that enforce the motion direction of robots along these paths. We will explain in more detail how such directional paths can be realized. If some clauses are nested in other clauses in the graph structure (for example, $c_2$ is nested inside $c_1$ due to the planarity requirement), additional paths are created to connect these clauses vertices, pointing from the inner clause to the outer clause ({\em e.g.}, the pink $(c_2, c_1)$ directional path). Finally, we create additional paths leading to the goals for the clause robots. These goal vertices are marked as $c_j^g$ in Fig.~\ref{figure:sketch}. We connect these paths to all clause vertices having a nesting level $0$ ($c_1, c_3$, and $c_4$) on the opposite side (see Fig.~\ref{figure:sketch}). 

Roughly, the reduction works as follows. If the \psat\, instance is satisfiable, then each clause robot can find a variable channel in the middle that will not have other clause robots using it from the opposite direction. This is true due to the monotone arrangement of the clauses. For example, the clause robot $r_{c_2}$, starting from vertex $c_2$, may choose to go to $x_2$, corresponding to setting $x_2 = true$. If $x_2 = true$, then $c_3$ cannot be true by setting $\neg x_2 = true$. Overall, for our example, we may set $x_1 = x_2 = x_4 = x_5 = true$ and $x_3 = false$. We can then let $r_{c_1}$ and $r_{c_2}$ go through variable channels for $x_1$ and $x_2$, respectively, in the downward direction. Similarly, we can let $r_{c_3}$ and $r_{c_4}$ go through variable channel for $x_3$, in the upward direction. Once these clause robots reach the other side of the variable channels, they can use the directional paths to reach their respective goals. Note that these paths are of appropriate lengths to make sure that the robots will not be delayed unnecessarily. For example, we will make robot $r_{c_4}$ spend a little more time to reach the variable channel for $x_3$ than it does for robot $r_{c_3}$. On the other hand, if a satisfiable assignment is not available, either a variable channel must be shared between both positive and negative clause robots or some clause robots must take a detour and use a variable channel of which the corresponding variable does not belong to the literals of that clause ({\em e.g.}, if $r_{c_2}$ goes to $c_1$ first and then $x_1$; $x_1$ does not appear in $c_2$). Both cases result in delays in the arrival of some robots. 

\subsection{Reducing Planar \sat\, to \pmtmpp: The Details}

We now provide the details leading to a complete proof of Theorem~\ref{t:time-npc}, starting from completing the \pmtmpp\, instance outlined in Fig.~\ref{figure:sketch}. The instances is constructed with a particular $k = 12m$ so that the \psat\, instance is satisfiable if and only if the corresponding \pmtmpp\, instance has an optimal solution with $12m$ time steps. First, we specify the lengths of different types of paths in Fig.~\ref{figure:sketch}.
\begin{enumerate}
\item {\em Variable channels}. A variable channel is a path of length $6m$ between $x_i$ and $\neg x_i$ (orange ones in Fig.~\ref{figure:sketch}). 
\item {\em Directional path from $c_j$ to a variable channel}. We call such a path a {\em forward path} from $c_j$. For a clause vertex $c_j$, the forward path from $c_j$ to $x_i$ or $\neg x_i$ (if one exists) has length $2j$. For example, the blue path from $c_1$ to $x_1, x_4$, and $x_5$ in Fig.~\ref{figure:sketch} all have length two, whereas the paths from $c_2$ to $x_2$ and $x_3$ are both of length four. 
\item {\em Directional path between clause vertices}. If a clause $c_{j_1}$ is directly nested inside $c_{j_2}$, we add a directional path from $c_{j_1}$ to $c_{j_2}$ and give it a length of two. These paths allow a clause robot to have a shortest path to its goal no matter which variable channel it goes through. For example, the pink path from $c_2$ to $c_1$ in Fig.~\ref{figure:sketch} is such a path, which allows $r_{c_3}$ to go through $\neg x_2$, $x_2$, $c_2$, $c_1$, and $c_4^g$ to reach $c_3^g$ (in $12m$ steps, as we establish shortly). 
\item {\em Directional path from a variable channel to $c_j$}. We call such a path a {\em backward} path to $c_j$. For a clause $c_j$ with a nesting level $\ell$, a backward path from $x_i$ or $\neg x_i$ to $c_j$ has a length of $2(m - \ell)$. 
For example, the path going from $x_2$ to $c_2$ has a length of $2(4 - 1) = 6$ and the path from $x_1$ to $c_1$ has a length of $2(4-0) = 8$. This ensures that paths like $x_1c_1$ and $x_2c_2c_1$ have the same length. 
\item {\em Goal paths for the clause robots}. The paths on which the goals of the clause robots are located (green paths in Fig.~\ref{figure:sketch}) have lengths such that the shortest path from from $c_j$ to $c_j^g$ has a length of $12m$, which can always be realized. Note that such constructions are not unique. However, the construction can be made unique, {\em e.g.}, by minimizing the number of added vertices. 
\end{enumerate}

\begin{figure}[ht]
\begin{center}
\includegraphics[width=0.60\textwidth]{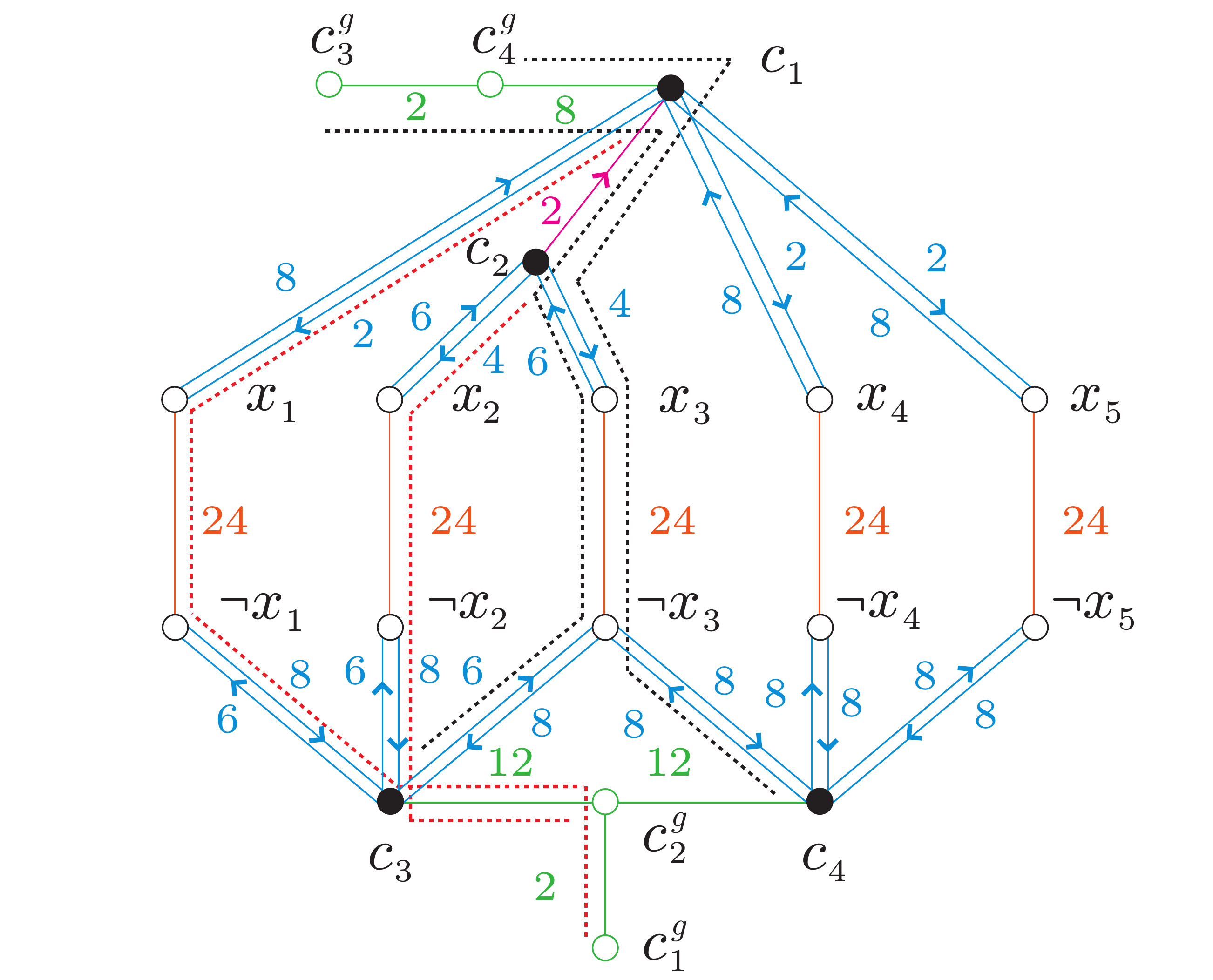} 
\end{center}
\caption{\label{figure:sketch-length} The skeleton of the \pmtmpp\, instance reduced from the \psat\, instance $(\{x_1, \ldots, x_5\}, \{c_1 = x_1 \vee x_4 \vee x_5, c_2 = x_2 \vee x_3, c_3 = \neg x_1 \vee \neg x_2 \neg x_3, c_4 = \neg x_3 \vee \neg x_4 \neg x_5\}))$, with path lengths added. Possible routes taken by the four clause robots following shortest paths are marked with red (downward) and black (upward) dotted lines.}
\end{figure}

For the example given in Fig.~\ref{figure:sketch}, the path lengths are added in Fig~\ref{figure:sketch-length}. It is straightforward to check that all the clause robots will require a minimum of $12m = 48$ steps to reach their respective goal vertices. 

The last main missing piece from the \pmtmpp\, instance is how to enforce the directional paths. There are three types of directional paths: forward paths from clause vertices, backward paths to clause vertices, and paths connecting two clause vertices. The method for enabling these directional paths are all similar; we do so by replacing each of these ``directional'' paths with a two-piece gadget, which is a simple, specialized sub-structure. We use examples to illustrate the gadget construction for each type of directional paths.

For the forward paths from clause vertices to variable channels, we only want to allow a clause robot to pass through the path at $t = 0$. For example, for the forward path from $c_3$ to $\neg x_1$ in Fig.~\ref{figure:sketch-length}, we only want to allow robot $r_{c_3}$ to pass through at time 
\begin{figure}[ht]
\begin{center}
\includegraphics[width=0.60\textwidth]{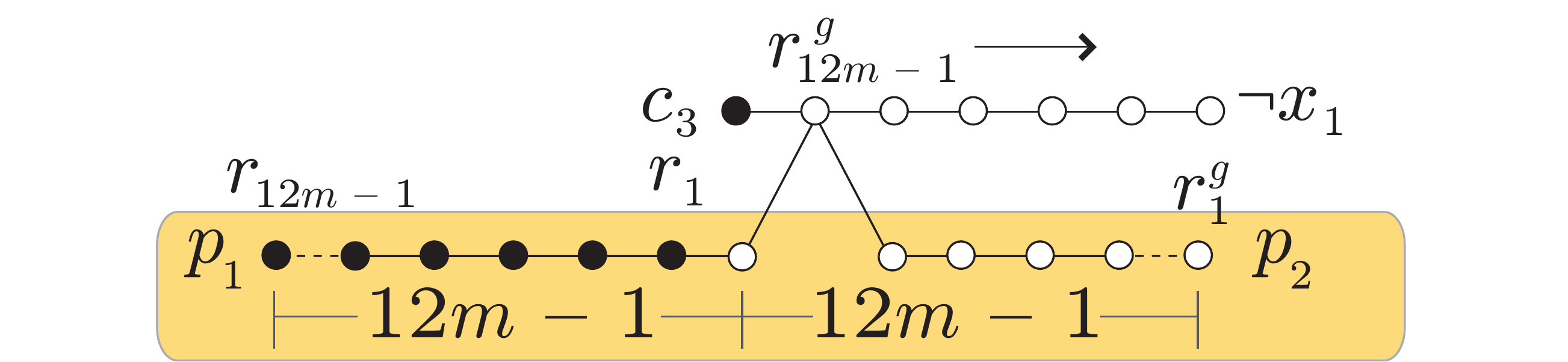} 
\end{center}
\caption{\label{figure:path-gadget} A gadget to enable unidirectional paths. The paths in the yellow shaded area are added to the path from $c_3$ to $\neg x_1$.}
\end{figure}
step $t = 0$ to time step $t = 6$. This can be enforced using the path gadget illustrated in Fig.~\ref{figure:path-gadget}, which also adds $12m-1$ additional robots. As shown in the figure, we attach a path ($p_1$) to the second vertex on the path from $c_3$ to $\neg x_1$ (in this case $r_{12m-1}^g$, which is also the goal vertex for robot $r_{12m-1}$). We then attach another path ($p_2$) at the same vertex. We put the $12m - 1$ robots on $p_1$ and require them to go to $p_2$. We make it so that each robot on $p_1$ requires $12m$ steps to reach its goal on $p_2$. The robots are placed on $p_1$ to allow desired clause robots to pass through. In the case of the path from $c_3$ to $\neg x_1$, no robot on $p_1$ will move to $r_{12m-1}^g$ at $t = 1$, thus allowing robot $r_{c_3}$ to go to $r_{12m-1}^g$ at $t = 1$. After $t = 1$, a continuous stream of robots on $p_1$ will move through $r_{12m - 1}^g$, forbidding any additional clause robots to move through the path connecting $c_3$ and $\neg x_1$. 

\begin{figure}[ht]
\begin{center}
\includegraphics[width=0.60\textwidth]{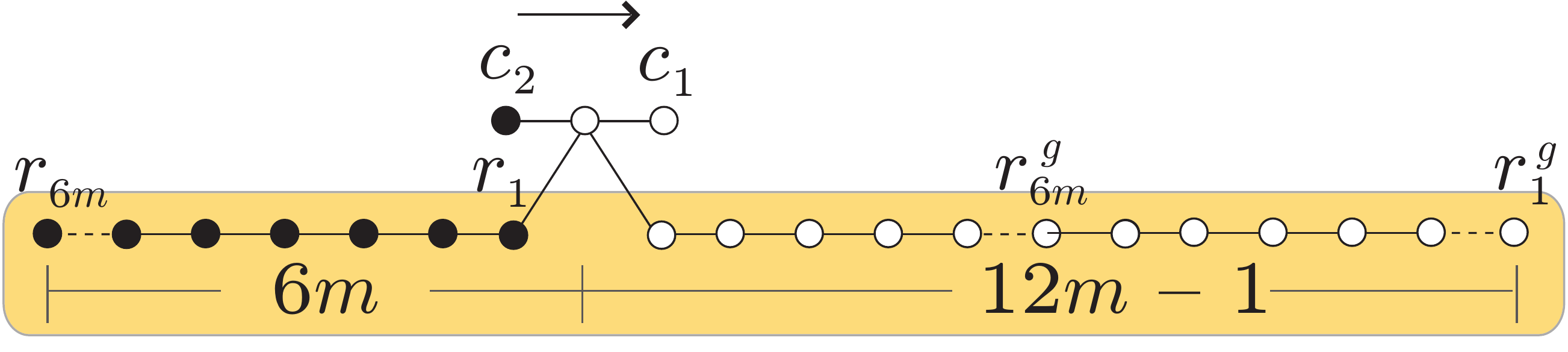} 
\end{center}
\caption{\label{figure:path-gadget-2} A gadget to enable unidirectional path from $c_2$ to $c_1$ that prevents traversing in the first $6m$ time steps.}
\end{figure}

We are left with two types of directional paths: backward paths to clause vertices and paths connecting two clause vertices. The gadget for these two types of directional paths are actually identical, since the utility of these paths is to allow a clause robot to move to its goal {\em after} passing through a variable channel. We use the directional path from $c_2$ to $c_1$ as an example. All we need to do is to forbid the path being used at all before $t = 6m$, thus preventing undesirable behaviors such as allowing $r_{c_2}$ to move from $c_2$ to $c_1$. We may do so simply by letting robots flowing through the path continuously for the first $6m$ steps (see Fig.~\ref{figure:path-gadget-2}). After that, the path becomes available for traversal in both directions. Nevertheless, such a path is effectively unidirectional because no clause robot can use it from the other direction ({\em i.e.}, from $c_1$ to $c_2$) when time optimality is enforced: if a robot uses such a path in the wrong direction after $6m$ steps, it cannot reach its goal in $12m$ steps. 

Finally, every variable channel is a simple straight line path of length $6m$ without any additional robot on it. To tally the number of vertices of the resulting instance, the skeleton graph has $O(mn)$ vertices ({\em e.g.}, Fig.~\ref{figure:sketch-length} can be decomposed into $2n$ paths between $c_3^g$ and $c_1^g$, with each such path having length $O(m)$). Then, there are no more than $O(m)$ forward and backward paths, and no more than $m$ paths between clause vertices. Since each of these three types of paths has $O(m)$ vertices, the entire construction has $O(mn + m^2)$ vertices and fewer robots, which can be done in polynomial time. Note that this analysis applies to all constructions in this paper.  

\begin{figure}[ht]
\begin{center}
\includegraphics[width=0.60\textwidth]{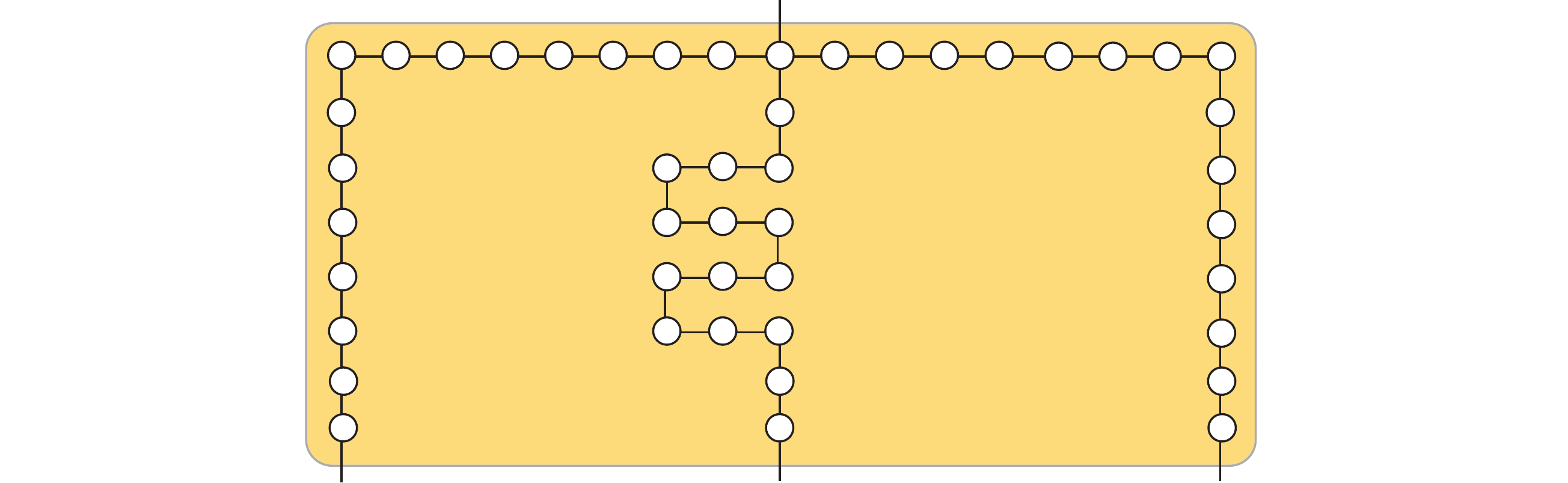} 
\end{center}
\caption{\label{figure:wrap} A possible arrangement of forward paths for a clause.}
\end{figure}

\remark It is straightforward to verify that the gadgets used in our construction do not change the planarity of the graph. Some readers may have concerns on the apparent difference of the edge lengths in the instance construction process--some edges seem longer than others whereas we assume every edge takes the same amount of time and distance to traverse. We note that edge lengths can always be made equal by scaling up the path lengths and bending certain paths. As an illustration, Fig.~\ref{figure:wrap} shows how forward paths for clauses (of length $16$ each) may be arranged while leaving space for backward paths and gadget paths. In general, linear scaling of path lengths creates quadratic amount of space for the accommodation of path length disparities.~\rqed

\noindent{\sc Proof of Theorem~\ref{t:time-npc}.} From the partially constructed \pmtmpp\, instance, we obtain a complete instance by setting $k = 12m$. If the \psat\, instance has a satisfiable assignment, then we can obtain collision free paths for all robots to simultaneously reach their goals in $12m$ time steps. To do so, we simply let a positive clause robot $r_{c_j}$ go from $c_j$ to some $x_i$ that is part of $c_j$ and is set to true in the assignment. We do the same for the negative clause robots, sending $r_{c_{j'}}$ to some $\neg x_{i'}$ that makes $c_{j'}$ true. Because a variable $x_i$ will not be set to true and false simultaneously in a satisfiable assignment, a positive clause robot and a negative clause robot will never reach the two ends of the same variable channel. All clause robots can then follow the directional paths to reach their goals in $12m$ steps. As mentioned earlier, in our example (Fig. \ref{figure:sketch-length}), we may set all variables but $x_3$ to true. We then let $r_{c_1}$ go to $x_1$, $r_{c_2}$ go to $x_2$, $r_{c_3}$ and $r_{c_4}$ go to $\neg x_3$ (the paths are illustrated in Fig. \ref{figure:sketch-length}). All robots will reach their respective goals at $t = 48$. 

On the other hand, if the constructed \pmpp\, instance has a solution requiring only $12m$ steps for all robots, then every single robot must start moving at $t = 0$, follow a shortest path, and never stop until the goal is reached. In particular, this means that a positive clause robot and a negative clause robot can never share the same variable channel. This is true because a positive (resp., negative) clause robot can reach some $x_i$ (resp., $\neg x_i$) at $0 < t \le 2m$. Since a variable channel has a length of $6m$, sharing it by a positive and a negative clause robot means that some robot must take more than $12m$ steps to reach its goal since the channel can only be used in one direction at a time. After it is used in one direction, more than $6m$ time steps has already passed. We then simply set a variable $x_i$ to be true (resp., false) if the corresponding channel is used by a positive (resp., negative) clause robot. Such an assignment is a satisfiable one for the original \psat\, problem. This proves that \pmtmpp\, is NP-hard.

To see that \pttmpp\, is NP-hard, we simply reuse the same construction but set $k$ to be $12m$ multiplied by the total number of robots. The rest of the proof remains the same. ~\qed

\remark As we know, the problem of planning time optimal trajectories when all robots are interchangeable is efficiently solvable \cite{YuLav13STAR}. By interchangeable, we mean that it is only required that all the goals are occupied by robots; it does not matter which robot occupies which goal. That is, there is a single {\em group} or {\em team} of robots. In the case of \pmpp, there are $n$ groups of robots (each group has a single robot). A natural question then arises: for how many groups of robots will optimal \pmpp\, problems become hard? It turns out two groups of robots are enough to make such problems computationally intractable. ~\rqed
\begin{theorem}\label{t:time-npc-2} \pttmpp\, and \pmtmpp\, remain NP-hard, even when there are only two groups of robots.\end{theorem}
\noindent{\sc Proof.} In the proof for Theorem~\ref{t:time-npc}, we group all positive clause robots and the associated robots on the path gadgets as one group.The rest of the robots form the other group. That is, referring to Fig. \ref{figure:sketch-length}, all robots above the set of variable channels belong to one group and all robots below belong to another. In our example, this means that $r_{c_1}$ and $r_{c_2}$ may go to either $c_1^g$ or $c_2^g$. Same is true for robots $r_{c_3}$ and $r_{c_4}$. It is straightforward to check that the robots for enforcing the directional paths (Fig. \ref{figure:path-gadget} and Fig. \ref{figure:path-gadget-2}) cannot be used for other purposes as every robot must continuously move $12m$ steps. Otherwise, time optimality will be affected. The proof of Theorem~\ref{t:time-npc} can then be applied to show that such formulations with two groups of robots remain NP-hard. ~\qed

\section{Intractability of Distance-Optimal Formulations}\label{section:distance}

\begin{figure}[ht]
\begin{center}
\includegraphics[width=0.60\textwidth]{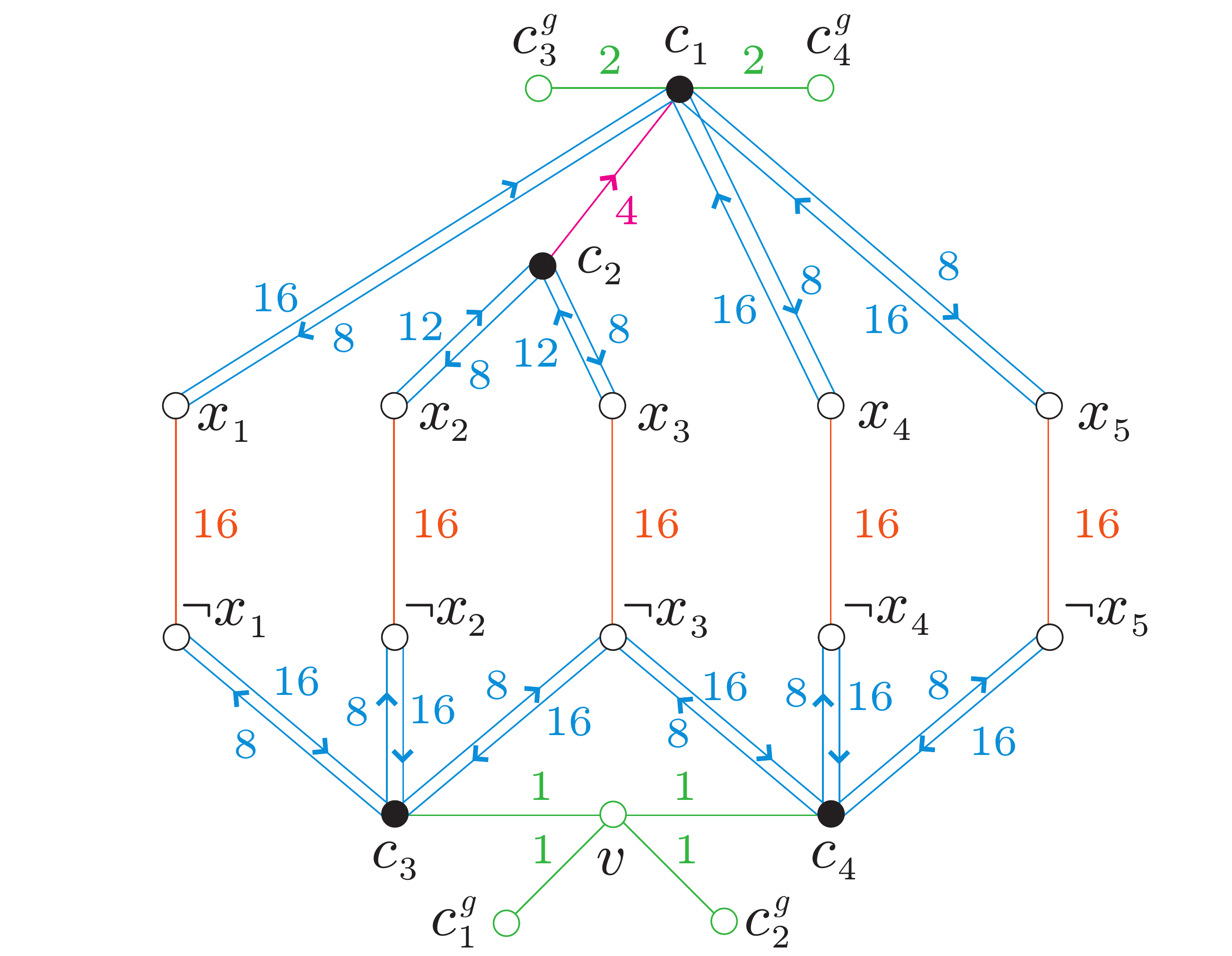} 
\end{center}
\caption{\label{figure:sketch-dist-length} The skeleton of the \ptdmpp\, instance reduced from the same \psat\, instance $(\{x_1, \ldots, x_5\}, \{c_1 = x_1 \vee x_4 \vee x_5, c_2 = x_2 \vee x_3, c_3 = \neg x_1 \vee \neg x_2 \neg x_3, c_4 = \neg x_3 \vee \neg x_4 \neg x_5\}))$, with path lengths added.}
\end{figure}

The intractability of distance-optimal formulations of \pmpp\, is more challenging to prove, owing to to the fact that it is not necessary to time-synchronize the paths of the robots, thus allowing more combinations of robot movements. Nevertheless, we again reduce from \psat\, and build distance optimal \pmpp\, instances, starting with \ptdmpp. The skeleton of the constructed \ptdmpp\, instance is shown in Fig.~\ref{figure:sketch-dist-length}, which has a structure similar to that of the converted \pmtmpp\, instance from Fig.~\ref{figure:sketch-length}. Like in Fig.~\ref{figure:sketch-length}, the lengths of the paths are given. The main difference at the skeleton level is that for distance optimality, we want to make sure that the shortest paths for all clause robots have not only the same length, but also in a sense ``parallel'' to each other. This is put in place to force a stronger form of synchronization among the robots. Again, we will show that the \psat\, instance is satisfiable if and only if the corresponding \ptdmpp\, has certain optimal solution cost. 

The path lengths are set as follows:
\begin{enumerate}
\item {\em Forward path from $c_j$}. All such paths have lengths of $2m$. This is 8 in our example. 
\item {\em Directional path between clause vertices}. If a clause $c_{j_1}$ is directly nested inside a clause vertex $c_{j_2}$, we add a path from $c_{j_1}$ to $c_{j_2}$ and make it have length 4. 
\item {\em Backward path to $c_j$}. Let the nesting level of $c_j$ be $\ell_j$, then such paths have lengths $4(m - \ell_j)$. For example, $x_2c_2$ has length $16 - 4 = 12$. This ensures that directional paths from all variable channels to a clause vertex of nesting level $0$ have the same length ({\em e.g.}, $x_1c_1$ and $x_2c_2c_1$), which is $4m$.
\item {\em Variable channels}. Each variable channel (orange paths in Fig.~\ref{figure:sketch}) has a length of $4m$. 
\item {\em Goal paths for the clause robots}. The length from any goal vertex for a clause robot to the closest clause vertex with a nesting level of $0$ is $2$. For example, $c_1c_3^g$ or $c_3c_1^g$. There are no additional structures on these paths. 
\end{enumerate}

Based on these path length settings, each clause robot requires a distance of at least $10m + 2$ to reach its goal. We now describe the details needed to specify the full \ptdmpp\, instance.  Since time synchronization is no longer helpful, gadgets such as those from Fig.~\ref{figure:path-gadget} and Fig.~\ref{figure:path-gadget-2} are no longer useful. Instead, we need a different method of enforcing directional paths that penalizes robot traveling in an undesirable direction. 

\begin{figure}[ht]
\begin{center}
\includegraphics[width=0.60\textwidth]{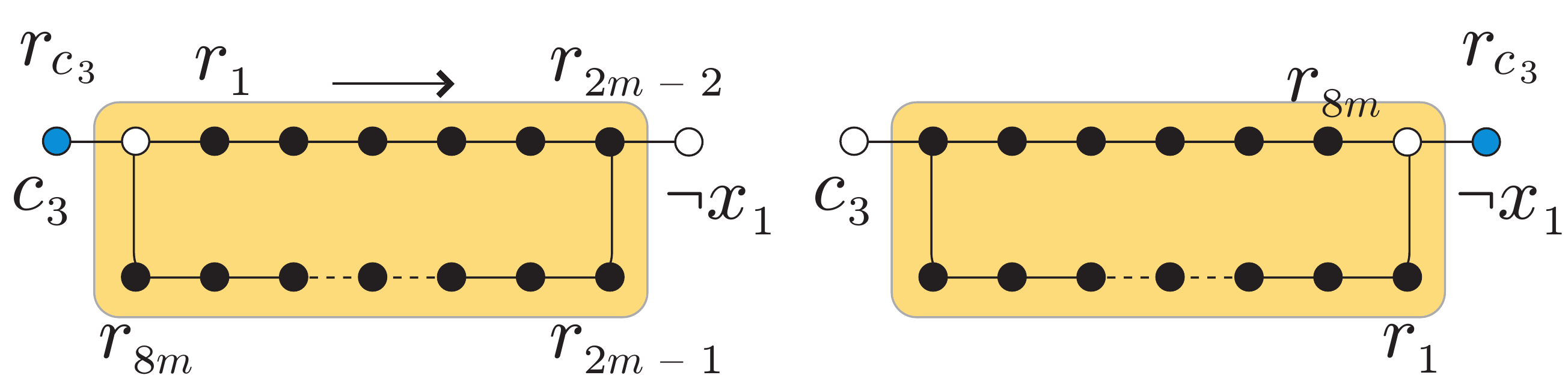} 
\end{center}
\caption{\label{figure:path-gadget-dist} A gadget to enable (total) distance optimal, unidirectional traversal from $c_3$ to $\neg x_1$. [left] The initial gadget configuration. [right] The target (goal) gadget configuration.}
\end{figure}

For a forward path from a clause vertex, for example from $c_3$ to $\neg x_1$, we construct the gadget illustrated in Fig.~\ref{figure:path-gadget-dist} to enforce unidirectional traversal. In the figure, the initial and final configurations of the gadget are shown in the left and the right picture (the shaded regions), respectively. Such a construction allows only clockwise rotation of the gadget cycle; rotating counterclockwise will incur a large distance penalty. Note that the added robots will incur more distance penalty if they travel outside the gadget structure. 

For backward paths to $c_j$, as well as the directional paths connecting $c_{j_1}$ to $c_{j_2}$, we let such a path be a simple straight line path. For example, the path $c_2c_1$ is a simple linear path of length $4$ with no additional robots in the middle. Note that a clause robot will never use such a path before crossing a variable channel since going through such a path adds the needed distance. For example, going from $c_1$ to $x_1$ along the forward path requires a distance of $8$ but going along the backward path requires a distance of $16$. 

\begin{figure}[ht]
\begin{center}
\includegraphics[width=0.60\textwidth]{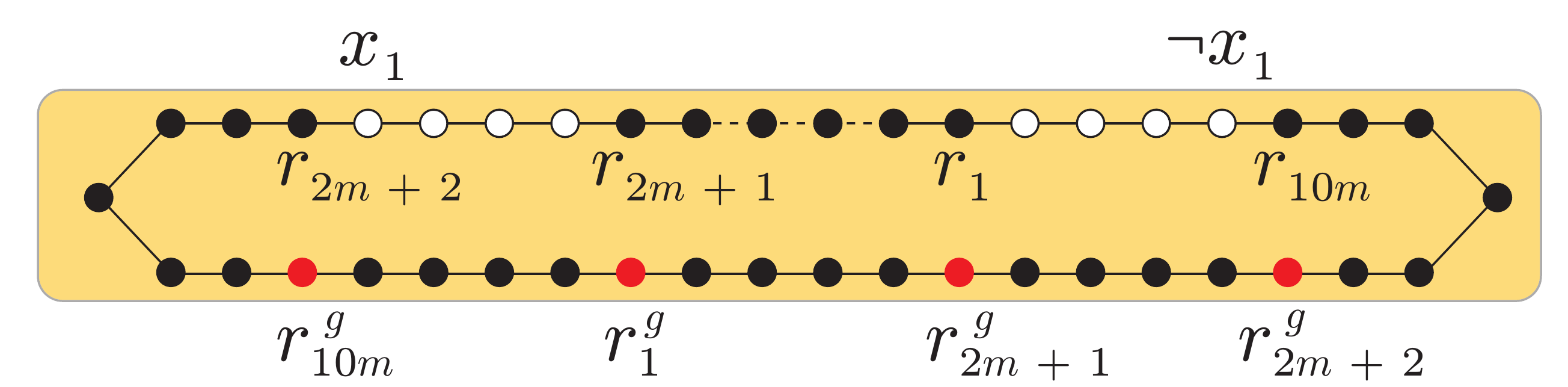} 
\end{center}
\caption{\label{figure:path-gadget-dist-2} The variable channel gadget for $x_1$ in the total distance optimal case. To ensure the robots on the gadget travel the shortest possible distance, all these robots may only move either in the same (clockwise or counterclockwise) direction.}
\end{figure}

Finally, a variable channel has the structure illustrated in Fig.~\ref{figure:path-gadget-dist-2}. The gadget contains a cycle of $12m$ vertices, with $10m$ additional robots on the cycle. For each of the $10m$ added robots, we require it to reach the diagonal location on the cycle. In particular, we marked the goals for robots $r_1, r_{2m+1}, r_{2m+2}$, and $r_{10m}$ in Fig.~\ref{figure:path-gadget-dist-2} (the red nodes). Given such a gadget, for all $10m$ robots to travel the minimum possible distance (which is $6m$ for each robot), the robots must either all rotate clockwise or all counterclockwise along the cycle. Effectively, such a gadget allows unidirectional traversal along the upper path between $x_1$ and $\neg x_1$, creating a unidirectional variable channel for clause robots. 

We now show that computing distance optimal solution for \pmpp\, problems is also NP-hard. 

\begin{theorem}\label{t:total-dist} \ptdmpp\, is NP-hard.\end{theorem}

\noindent{\sc Proof.} Given the \pmpp\, instance that has been constructed, we obtain a complete \ptdmpp\, instance by setting $k$ to be the sum of the minimum possible distance for all the robots. Assuming the \psat\, instance has a satisfiable assignment, we let a clause robot move to a variable channel whose end literal makes that clause true. In our example, we may again let all variables but $x_3$ be true. We then let $r_{c_1}$ go to $x_1$, $r_{c_2}$ go to $x_2$, and both $r_{c_3}$ and $r_{c_4}$ go to $\neg x_3$. There robots can then pass the variable channel to reach their respective goals. It is clear that all robots may do so following a shortest possible path. In particular, a variable channel can transfer up to $m$ clause robots in one direction optimally. Also, we note that the clause robots only need to synchronize their movements at variable channels. Otherwise, the order of their movements do not affect distance optimality. For example, the movement of $r_{c_1}$ and $r_{c_2}$ can be mostly decoupled; they only share one edge in their respective shortest paths (the edge $(c_3, v)$ in Fig.~\ref{figure:sketch-dist-length}). 

On the other hand, if the \ptdmpp\, instance has a distance optimal solution of $k$ total steps, each robot must follow a shortest possible path. For a clause robot $r_{c_j}$ to follow a shortest path, it must first travel along a forward path gadget that go from $c_j$ to a variable channel, in $2m$ steps. Because a variable channel can only be used in one direction to optimally transfer clause robots, the distance optimal solution partitions the variable channels into two sets based on the travel direction of the clause robots. Similar to the time-optimal case, based on this partition of variable channels, a satisfiable assignment for the corresponding \psat\, instance can then be easily extracted. ~\qed

\begin{theorem}\label{t:max-dist} \pmdmpp\, is NP-hard.\end{theorem}

\begin{figure}[ht]
\begin{center}
\includegraphics[width=0.60\textwidth]{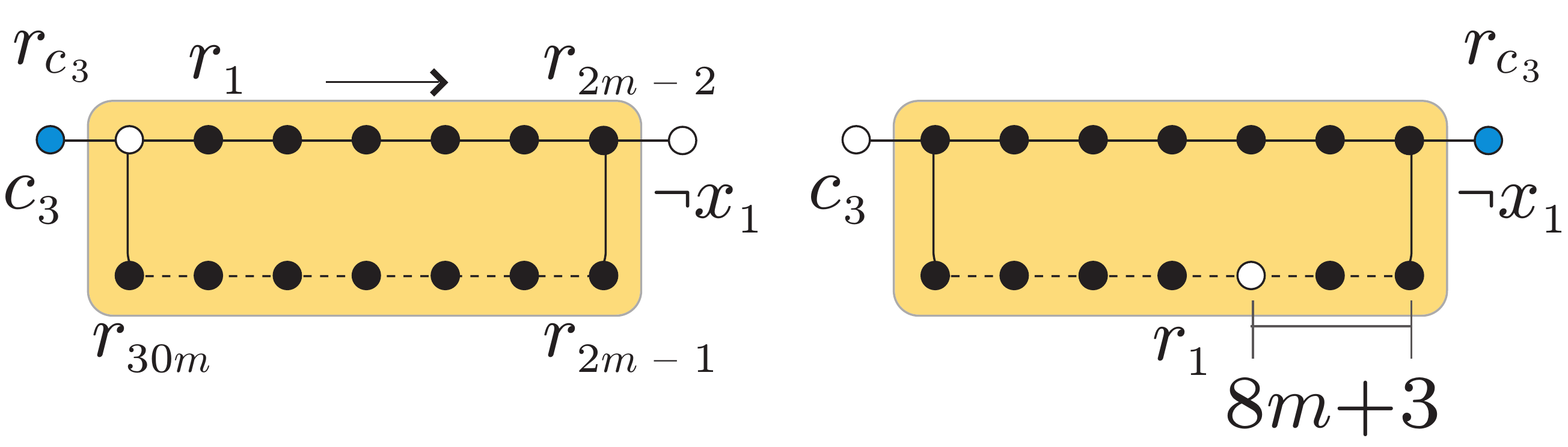} 
\end{center}
\caption{\label{figure:path-gadget-dist-l} A gadget to enable distance optimal, unidirectional traversal from $c_3$ to $\neg x_1$ while ensuring the added robots move at least $10m + 2$ steps. [left] The initial gadget configuration. [right] The target (goal) gadget configuration.}
\end{figure}

\noindent{\sc Proof.} In our construction of the \ptdmpp\, instance, there are three different shortest distances for the robots that are involved. A clause robot needs to travel $10m + 2$ steps, a robot added to a directional path gadget needs to travel $2m - 2$ steps, and a robot added to the variable gadget needs to travel $6m$ steps. However, we can easily change the gadgets such that all the involved robots need to travel $10m + 2$ steps. For the forward path gadget pictured in Fig.~\ref{figure:path-gadget-dist}, we may do so by making the lower path of the cycle much larger ({\em e.g.}, to have about $30m$ vertices and robots, see Fig.~\ref{figure:path-gadget-dist-l}) to again ensure unidirectional traversal through the gadget. 
For the variable channel gadget, we may simply enlarge the cycle to have $20m + 4$ vertices, add more robots, and again make all the added robots go to diagonal locations. After the gadget modifications are complete, we obtain a complete \pmdmpp\, instance by setting $k = 10m + 2$. The rest of the proof is then identical to that for \ptdmpp. ~\qed

\section{Conclusion and Discussion}\label{section:conclusion}
In this paper, we have established the NP-hardness of four common versions of optimal multi-robot path planning problems on planar graphs. Because these problems are clearly in NP, we may further conclude that they are NP-complete, {\em i.e.}, 

\begin{theorem}\label{t:all} \pttmpp, \pmtmpp, \ptdmpp, and \pmdmpp\, are all NP-complete problems.\end{theorem}

Theorem~\ref{t:all} implies that optimal planning and control of multiple robots in a discrete planar environment, such as coordinating many automobiles on a planar road network to minimize delay, is a computationally intractable task. In particular, the difficulty appears to arise when two or more groups of robots want to move in opposite directions through the same set of channels, thus creating resource contention among the robots. From a practical standpoint, our observation suggests that the best approach towards solving such problems, as the number of robots becomes large, is perhaps to engineer the environment carefully to minimize path sharing among the robots. Alternatively, an equally interesting open question is whether optimal \mpp\, and \pmpp\, problems are APX-hard or they admit polynomial time approximation schemes (PTAS). 


\bibliographystyle{IEEEtranN}
\bibliography{ral-15-200}

\begin{thebibliography}{26}
\providecommand{\natexlab}[1]{#1}
\providecommand{\url}[1]{#1}
\csname url@samestyle\endcsname
\providecommand{\newblock}{\relax}
\providecommand{\bibinfo}[2]{#2}
\providecommand{\BIBentrySTDinterwordspacing}{\spaceskip=0pt\relax}
\providecommand{\BIBentryALTinterwordstretchfactor}{4}
\providecommand{\BIBentryALTinterwordspacing}{\spaceskip=\fontdimen2\font plus
\BIBentryALTinterwordstretchfactor\fontdimen3\font minus
  \fontdimen4\font\relax}
\providecommand{\BIBforeignlanguage}[2]{{%
\expandafter\ifx\csname l@#1\endcsname\relax
\typeout{** WARNING: IEEEtranN.bst: No hyphenation pattern has been}%
\typeout{** loaded for the language `#1'. Using the pattern for}%
\typeout{** the default language instead.}%
\else
\language=\csname l@#1\endcsname
\fi
#2}}
\providecommand{\BIBdecl}{\relax}
\BIBdecl

\bibitem[B\"ohringer(2004)]{Boh04}
K.~F. B\"ohringer, ``Towards optimal strategies for moving droplets in digital
  microfluidic systems,'' in \emph{Proceedings IEEE International Conference on
  Robotics \& Automation}, 2004.

\bibitem[Grocholsky et~al.(2006)Grocholsky, Keller, Kumar, and
  Pappas]{GroKelKumPap06}
B.~Grocholsky, J.~Keller, V.~Kumar, and G.~Pappas, ``Cooperative air and ground
  surveillance,'' \emph{IEEE Robotics and Automation Magazine}, vol.~13, no.~3,
  pp. 16--25, Sep 2006.

\bibitem[Wurman et~al.(2008)Wurman, D'Andrea, and Mountz]{WurDanMou08}
P.~R. Wurman, R.~D'Andrea, and M.~Mountz, ``Coordinating hundreds of
  cooperative, autonomous vehicles in warehouses,'' \emph{AI Magazine},
  vol.~29, no.~1, pp. 9--19, 2008.

\bibitem[Knepper et~al.(2013)Knepper, Layton, Romanishin, and
  Rus]{KneLayRomRus13}
R.~Knepper, T.~Layton, J.~Romanishin, and D.~Rus, ``Ikeabot: An autonomous
  multi-robot coordinated furniture assembly system,'' in \emph{Proceedings
  IEEE International Conference on Robotics \& Automation}, 2013, pp. 855--862.

\bibitem[Roozbehani and D'Andrea(2011)]{RooRaf11}
H.~Roozbehani and R.~D'Andrea, ``Adaptive highways on a grid,'' in
  \emph{Robotics Research}.\hskip 1em plus 0.5em minus 0.4em\relax Springer,
  2011, pp. 661--680.

\bibitem[Story(1879)]{Sto1879}
E.~W. Story, ``Note on the `15' puzzle,'' \emph{American Journal of
  Mathematics}, vol.~2, pp. 399--404, 1879.

\bibitem[Wilson(1974)]{Wil74}
R.~M. Wilson, ``Graph puzzles, homotopy, and the alternating group,''
  \emph{Journal of Combinatorial Theory (B)}, vol.~16, pp. 86--96, 1974.

\bibitem[Loyd(1959)]{Loy59}
S.~Loyd, \emph{Mathematical Puzzles of Sam Loyd}.\hskip 1em plus 0.5em minus
  0.4em\relax New York: Dover, 1959.

\bibitem[Kornhauser et~al.(1984)Kornhauser, Miller, and Spirakis]{KorMilSpi84}
D.~Kornhauser, G.~Miller, and P.~Spirakis, ``Coordinating pebble motion on
  graphs, the diameter of permutation groups, and applications,'' in
  \emph{Proceedings IEEE Symposium on Foundations of Computer Science}, 1984,
  pp. 241--250.

\bibitem[Ryan(2008)]{Rya08}
M.~R.~K. Ryan, ``Exploiting subgraph structure in multi-robot path planning,''
  \emph{Journal of Artificial Intelligence Research}, vol.~31, pp. 497--542,
  2008.

\bibitem[Standley and Korf(2011)]{StaKor11}
T.~Standley and R.~Korf, ``Complete algorithms for cooperative pathfinding
  problems,'' in \emph{Proceedings International Joint Conference on Artificial
  Intelligence}, 2011, pp. 668--673.

\bibitem[Wagner and Choset(2011)]{WagChoC11}
G.~Wagner and H.~Choset, ``M*: A complete multirobot path planning algorithm
  with performance bounds,'' in \emph{Proceedings IEEE/RSJ International
  Conference on Intelligent Robots \& Systems}, 2011, pp. 3260--3267.

\bibitem[Surynek(2012)]{Sur12}
P.~Surynek, ``Towards optimal cooperative path planning in hard setups through
  satisfiability solving,'' in \emph{Proceedings 12th Pacific Rim International
  Conference on Artificial Intelligence}, 2012.

\bibitem[Goldreich(1984)]{Gol84}
O.~Goldreich, ``Finding the shortest move-sequence in the graph-generalized
  15-puzzle is {NP}-hard,'' 1984, laboratory for Computer Science,
  Massachusetts Institute of Technology, {U}npublished manuscript.

\bibitem[Ratner and Warmuth(1990)]{RatWar90}
D.~Ratner and M.~Warmuth, ``The $(n^2-1)$-puzzle and related relocation
  problems,'' \emph{Journal of Symbolic Computation}, vol.~10, pp. 111--137,
  1990.

\bibitem[Surynek(2010)]{Sur10}
P.~Surynek, ``An optimization variant of multi-robot path planning is
  intractable,'' in \emph{Proceedings AAAI National Conference on Artificial
  Intelligence}, 2010, pp. 1261--1263.

\bibitem[Yu and LaValle(2013{\natexlab{a}})]{YuLav13AAAI}
J.~Yu and S.~M. LaValle, ``Structure and intractability of optimal multi-robot
  path planning on graphs,'' in \emph{Proceedings AAAI National Conference on
  Artificial Intelligence}, 2013, pp. 1444--1449.

\bibitem[Hopcroft et~al.(1984)Hopcroft, Schwartz, and Sharir]{HopSchSha84}
J.~E. Hopcroft, J.~T. Schwartz, and M.~Sharir, ``On the complexity of motion
  planning for multiple independent objects; {PSPACE}-hardness of the
  ``warehouseman's problem'','' \emph{The International Journal of Robotics
  Research}, vol.~3, no.~4, pp. 76--88, 1984.

\bibitem[Spirakis and Yap(1984)]{SpiYak84}
P.~Spirakis and C.~K. Yap, ``Strong {NP}-hardness of moving many discs,''
  \emph{Information Processing Letters}, vol.~19, no.~1, pp. 55--59, 1984.

\bibitem[Solovey and Halperin(2015)]{SolHal15}
K.~Solovey and D.~Halperin, ``On the hardness of unlabeled multi-robot motion
  planning,'' in \emph{Robotics: Science and Systems (RSS)}, 2015.

\bibitem[Hearn and Demaine(2005)]{HeaDem05}
R.~A. Hearn and E.~D. Demaine, ``{PSPACE}-completeness of sliding-block puzzles
  and other problems through the nondeterministic constraint logic model of
  computation,'' \emph{Theoretical Computer Science}, vol. 343, no.~1, pp.
  72--96, 2005.

\bibitem[Yu and Rus(2014)]{YuRus14WAFR}
J.~Yu and D.~Rus, ``Pebble motion on graphs with rotations: Efficient
  feasibility tests and planning,'' in \emph{Proceedings Workshop on
  Algorithmic Foundations of Robotics}, 2014.

\bibitem[Dijkstra(1959)]{dijkstra1959note}
E.~W. Dijkstra, ``A note on two problems in connexion with graphs,''
  \emph{Numerische mathematik}, vol.~1, no.~1, pp. 269--271, 1959.

\bibitem[Garey and Johnson(1979)]{GarJoh79}
M.~R. Garey and D.~S. Johnson, \emph{Computers and Intractability: A Guide to
  the Theory of NP-Completeness}.\hskip 1em plus 0.5em minus 0.4em\relax W. H.
  Freeman, 1979.

\bibitem[de~Berg and Khosravi(2010)]{BerKho10}
M.~de~Berg and A.~Khosravi, ``Optimal binary space partitions in the plane,''
  in \emph{Computing and Combinatorics}.\hskip 1em plus 0.5em minus 0.4em\relax
  Springer, 2010, pp. 216--225.

\bibitem[Yu and LaValle(2013{\natexlab{b}})]{YuLav13STAR}
J.~Yu and S.~M. LaValle, ``Multi-agent path planning and network flow,'' in
  \emph{Algorithmic Foundations of Robotics {X}, Springer Tracts in Advanced
  Robotics}.\hskip 1em plus 0.5em minus 0.4em\relax Springer Berlin/Heidelberg,
  2013, vol.~86, pp. 157--173.

\end{thebibliography}

\end{document}